%% file: root.tex
\documentclass[letterpaper, 10 pt, conference]{ieeeconf}  

\IEEEoverridecommandlockouts                              

\usepackage{amsmath,amsfonts}
\usepackage{amssymb}

\usepackage{algorithm}
\usepackage[noend]{algorithmic}
\usepackage{color}

\usepackage{array}
\usepackage{textcomp}
\usepackage{stfloats}

\usepackage{verbatim}
\usepackage{graphicx}
\usepackage{amsmath}
\usepackage{amsfonts}
\usepackage{amssymb}
\let\labelindent\relax
\usepackage{enumitem}
\usepackage{mathtools}
\usepackage{adjustbox}
\usepackage{cite}
\usepackage[utf8]{inputenc} 
\usepackage[T1]{fontenc}    
\usepackage{hyperref}       
\usepackage{url}            
\usepackage{booktabs}       
\usepackage{amsfonts}       
\usepackage{nicefrac}       
\usepackage{microtype}      
\usepackage{lipsum}
\usepackage{fancyhdr}       
\usepackage{graphicx}       
\usepackage{wrapfig}
\usepackage{array}
\usepackage{multirow}
\usepackage{threeparttable}
\usepackage{subcaption}
\usepackage{adjustbox}
\usepackage{multirow}

\title{\LARGE \bf
VFP: Variational Flow-Matching Policy for \\ Multi-Modal Robot Manipulation
}

\author{Xuanran Zhai$^{1}$, Qianyou Zhao$^{2}$, Qiaojun Yu$^{3}$, Ce Hao$^{*1}$ \\ $^{1}$National University of Singapore $^{2}$Shanghai Jiao Tong University $^{3}$Shanghai AI Lab \quad\thanks{$^{*}$ Corresponding author: cehao@u.nus.edu}}

\begin{document}

\maketitle


\input{Sections/0_abstract}
\input{Sections/1_Introduction}

\input{Sections/2_Related_Works}
\input{Sections/3_Preliminary}

\input{Sections/4_Method}

\input{Sections/5_Experiment}

\input{Sections/6_Conclusion}

\bibliographystyle{IEEEtran}
\bibliography{references}

\end{document}

%% file: Sections/0_abstract.tex
\begin{abstract}
Flow-matching-based policies have recently emerged as a promising approach for learning-based robot manipulation, offering significant acceleration in action sampling compared to diffusion-based policies. However, conventional flow-matching methods struggle with multi-modality, often collapsing to averaged or ambiguous behaviors in complex manipulation tasks. To address this, we propose the Variational Flow-Matching Policy (VFP), which introduces a variational latent prior for mode-aware action generation and effectively captures both task-level and trajectory-level multi-modality. VFP further incorporates Kantorovich Optimal Transport (K-OT) for distribution-level alignment and utilizes a Mixture-of-Experts (MoE) decoder for mode specialization and efficient inference. We comprehensively evaluate VFP on $41$ simulated tasks and $3$ real-robot tasks, demonstrating its effectiveness and sampling efficiency in both simulated and real-world settings. Results show that VFP achieves a $49\%$ relative improvement in task success rate over standard flow-based baselines in simulation, and further outperforms them on real-robot tasks, while still maintaining fast inference and a compact model size. More details are available on our project page: \url{https://sites.google.com/view/varfp/}
\end{abstract}

%% file: Sections/1_Introduction.tex
\section{Introduction} \label{Sec: intro}

Learning-based methods have driven major advances in robot manipulation, enabling robots to generalize across a wide range of tasks from expert demonstrations~\cite{wang2023mimicplaylonghorizonimitationlearning, handa2019dexpilotvisionbasedteleoperation,shridhar2022perceiveractormultitasktransformerrobotic}. Among these approaches, diffusion-based policies have shown impressive results in multi-modal imitation learning by leveraging stochastic denoising to capture the diverse action distributions present in real-world tasks~\cite{diffplcy,3ddiff, wang2024sparsediffusionpolicysparse, lu2024manicm, wang2024onestepdiffusionpolicyfast}. Yet, the slow, iterative sampling process of diffusion models limits their practicality for real-time control. To address this, flow-matching (FM) policies~\cite{rouxel2024flowmatchingimitationlearning, flowpolicy, zhang2025affordancebasedrobotmanipulationflow, braun2024riemannianflowmatchingpolicy} have emerged as a promising alternative, employing probabilistic flow-based ODEs for much faster action generation—requiring only a single ODE integration step. In practice, flow-matching policies can reduce sampling time to just 20\% of that needed by diffusion models, making them highly attractive for time-sensitive robot manipulation tasks.

\begin{figure}[t]
    \centering
    \includegraphics[width=\linewidth]{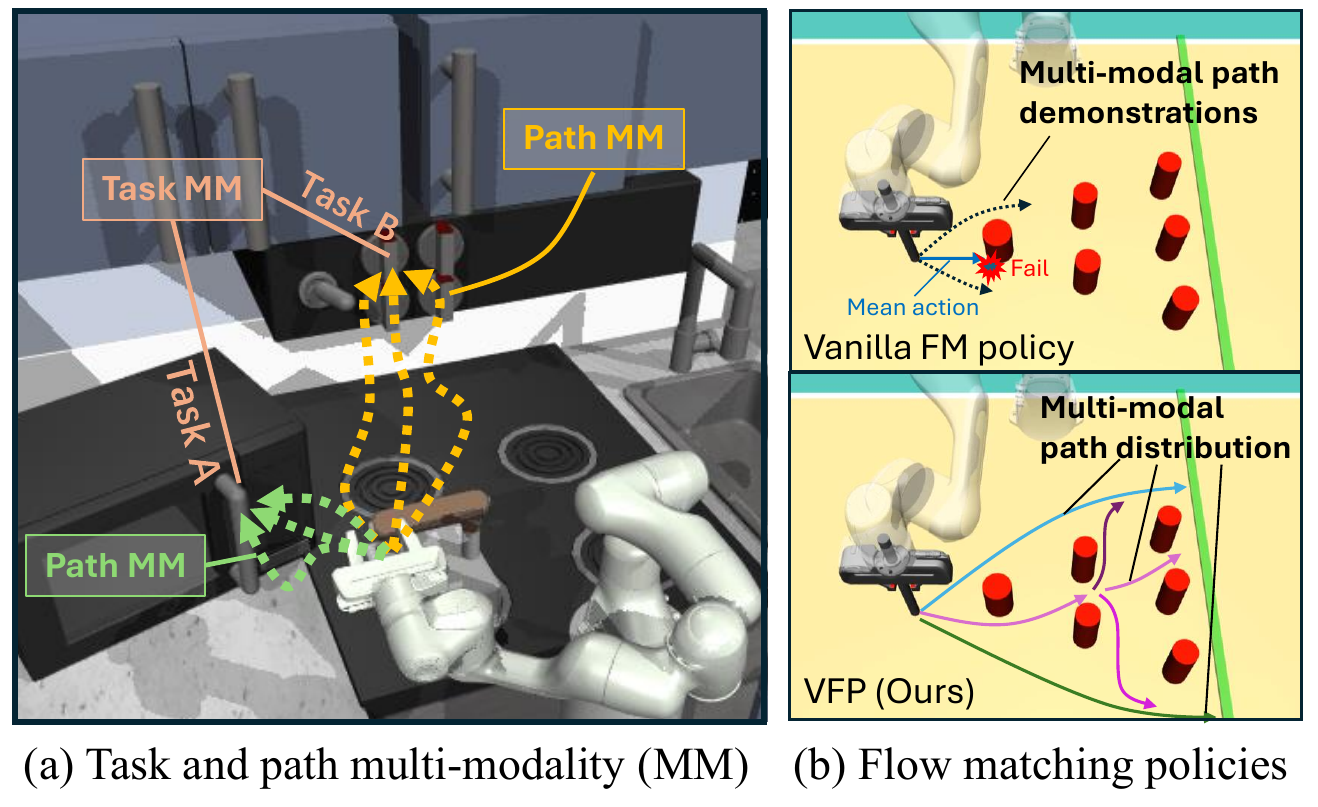}
        \caption{\textbf{(a)} In Franka Kitchen environment, the robot finishes multiple tasks and each task has different paths, which requires the policy to capture the task and path multi-modality (MM). \textbf{(b)} In the avoiding task, the vanilla flow-matching (FM) policy learns the mean action over all demonstrations, while our variational flow-matching policy (VFP) successfully follows the MM path distribution.}
    \label{Fig: teaser}
\vspace{-6mm}
\end{figure}

However, flow-matching policies struggle to capture the inherent multi-modal distributions found in robot manipulation tasks. As illustrated in Fig.~\ref{Fig: teaser} (a), a robot operating in a kitchen environment faces both task-level and path-level multi-modality, where it must not only accomplish different objectives but may also follow various demonstrated paths to complete each task. Flow matching, formulated as an ODE-based generative model, is fundamentally limited in representing such multi-modal behavior with neural networks~\cite{samaddar2025efficientflowmatchingusing, chen2025gaussianmixtureflowmatching,guo2025variationalrectifiedflowmatching}. For example, as shown in Fig.~\ref{Fig: teaser} (b), when a robot must plan collision-free paths, the flow-matching policy often averages over all demonstrated options, resulting in actions that do not correspond to any valid path. This averaging effect leads to significant failures in imitation learning for manipulation tasks where representing diverse behaviors is essential.

To address this limitation, we propose the \textit{Variational Flow-Matching Policy} (VFP) for fast and expressive multi-modal policy generation in robot manipulation (Fig.~\ref{Fig: teaser} (b)). The key idea of VFP is to use a variational latent prior to capture the inherent multi-modality in expert demonstrations, while employing a flow-matching model as a conditional decoder that generates actions specific to each mode. This variational formulation shifts the burden of mode identification to the latent prior, effectively avoiding the averaging effect of standard flow matching and enabling the policy to model diverse behaviors. To further align the learned action distribution with the multi-modal expert data, we introduce Kantorovich Optimal Transport (K-OT)~\cite{ot1,ot2,ot3} as a distribution-level regularization, which directly matches the predicted and expert action distributions for each state and promotes clear mode separation. Additionally, we implement the flow decoder as a Mixture-of-Experts (MoE) model~\cite{moe1,moe2}, allowing each expert to focus on a distinct behavioral mode and improving both multi-modality representation and sampling efficiency. Through the integration of variational inference, K-OT distribution matching, and MoE expert specialization, VFP achieves robust and real-time multi-modal imitation learning in complex robot manipulation tasks.






We validate the effectiveness and computational efficiency of VFP across both simulated and real-robot environments. 
In simulation, we evaluate on the Franka Kitchen benchmark~\cite{frankaKitchen} for task multi-modality, where multiple tasks coexist and can be completed in arbitrary order. In this setting, FlowPolicy~\cite{flowpolicy} exhibits unstable, indecisive behaviors due to mode averaging, while VFP clearly distinguishes and executes distinct task modes, achieving an $11.5\%$ improvement.
It also surpasses FlowPolicy on path multi-modality benchmarks including D3IL~\cite{d3il}, Adroit~\cite{adroit}, and Meta-World~\cite{metaworld}, with a $61.7\%$ average improvement across 30 tasks. 
In ablation studies, our results show that VFP effectively balances multi-modal policy representation with computational efficiency, consistently achieving high performance in complex tasks without sacrificing the fast inference advantage of flow-matching models. Beyond simulation, we further evaluate on 3 real-robot tasks, where VFP achieves higher success counts than both DP and FlowPolicy, confirming that its advantage transfers to the real world.

In summary, our main contributions are:
\begin{itemize}
    \item We propose Variational Flow-Matching Policy (VFP), a new imitation learning framework that uses a variational latent prior for mode identification, effectively overcoming the averaging problem in flow-matching models.
    \item We introduce Kantorovich Optimal Transport (K-OT) for explicit distribution alignment and employ a Mixture-of-Experts (MoE) decoder to enhance multi-modal expressiveness and computational efficiency.
    \item We extensively validate VFP on diverse simulation benchmarks and 3 real-robot tasks, achieving a $49\%$ average improvement in success rate, up to $94\%$ on highly multi-modal settings, and consistently higher success counts on real robots, while maintaining fast inference—a key advantage over stochastic methods.
\end{itemize}

%% file: Sections/2_Related_Works.tex
\section{Related Works}

%



\paragraph{\textbf{Flow Matching}}
Flow Matching (FM)~\cite{fmLipman,rectifiedFlow} is a class of generative methods that learn continuous-time vector fields to transport samples from a source distribution to a target distribution. The roots of FM can be traced back to Normalizing Flows (NF)~\cite{normFlow1,normFlow2,normFlow3} and Neural ODEs~\cite{neuralODE}, where invertible transformations or flow-based dynamics are used to model complex distributions. In contrast to traditional NF, FM directly learns a velocity field without requiring invertibility, enabling more flexible modeling.
Recent advances in Flow Matching include several instantiations of the framework, such as Rectified Flow~\cite{rectifiedFlow}, Consistency Models~\cite{CFM}, and Flow Matching for Score-based Modeling~\cite{fmLipman}, which propose different formulations of matching objectives. These methods have shown competitive generation quality.


\paragraph{\textbf{Multi-modality in Robot Manipulation}}
Multi-modality (MM), a prevalent characteristic in robotic manipulation tasks, primarily consists of task and path multi-modality. Task MM arises in environments with multiple valid task objectives, whereas path MM is even more common, as there are typically multiple feasible trajectories to achieve the same goal. For example, an object can be pushed from one location to another along different paths. To model such diversity, prior works have explored mixture models like MDNs~\cite{bishop1994mixture}, and latent-variable methods such as conditional VAEs~\cite{hausman2017multimodalimitationlearningunstructured}. More recently, diffusion-based approaches~\cite{diffplcy, 3ddiff} have shown success in capturing multi-modal distributions through denoising-based generation. However, while flow matching holds great potential for imitation learning due to its fast inference speed, apart from some recent works~\cite{zhang2025affordancebasedrobotmanipulationflow} that rely on large pretrained vision-language models, to the best of our knowledge, no existing flow-based approach has addressed multi-modal behavior modeling in imitation learning.

%% file: Sections/3_Preliminary.tex
\section{Preliminary} \label{sec:prelim}

\paragraph{\textbf{Problem Formulation}}
We consider robot manipulation tasks $\mathcal{T}$, where the goal is to train a policy $p_\theta(a|s)$ parameterized by $\theta$ via imitation learning. States $s \in \mathcal{S}$ and actions $a \in \mathcal{A}$ are observed and executed in a manipulation environment. Given a dataset of expert demonstrations $\mathcal{D} = \{(s_i, a_i)\}$, the policy is trained by maximizing the conditional log-likelihood $\mathbb{E}_{(s,a) \sim \mathcal{D}} \left[ \log\, p(a \mid s) \right]$.

\paragraph{\textbf{Flow Matching}}
Flow matching formulates generative modeling as learning a velocity field that deterministically transports samples from a source distribution to a target distribution along a continuous trajectory. In the context of policy learning for robot manipulation, this corresponds to learning a time-dependent velocity field $v_\theta(a, t, s)$ that maps an initial action $a_0 \sim p_0(a_0 \mid s)$ to a target action $a_1 \sim p_1(a_1 \mid s)$.

At inference time, a sample $a_0$ is drawn from the source distribution $p_0(a_0 \mid s)$ and treated as the initial condition for an ordinary differential equation (ODE). This ODE is solved by integrating the velocity field $v_\theta(a_t, t, s)$ forward from $t=0$ to $t=1$, with $a_t$ following a trajectory (e.g., linear interpolation: $a_t = (1-t)a_0 + t a_1$ in rectified flow matching~\cite{rectifiedFlow}) between $a_0$ and $a_1$. The likelihood of a data point $a_1$ under this model can be assessed using the instantaneous change-of-variables formula:
\vspace{-1mm}
\begin{equation} \label{Eqn: FM integral}
    \log p_1(a_1) = \log p_0(a_0) + \int_1^0 \mathrm{div}\, v_\theta(a_t, t, s)\, dt,
\end{equation}
where $\mathrm{div}$ denotes the divergence operator. 
During training, pairs of samples $(a_0, a_1)$ and a time $t \in [0,1]$ are drawn, and the interpolated location $a_t = (1-t)a_0 + t a_1$ is computed. The ground-truth velocity at this point is $v(a_0, a_1, t) = a_1 - a_0$, which is used as the target for the model's parametric velocity field. The flow matching objective is to minimize the mean squared error between the learned and ground-truth velocities:
\begin{equation}
    \mathcal{L}_\mathrm{FM} = \mathbb{E}_{s, a_0, a_1, t} \left[ \| v_\theta(a_t, t, s) - (a_1 - a_0) \|^2 \right].
\end{equation}
This approach allows flow matching to flexibly model complex action distributions by learning deterministic trajectories between source and target actions for any given state $s$. The method provides efficient sampling, making it well-suited for robot learning, especially in real-world settings.

\paragraph{\textbf{Velocity Ambiguity of Flow Matching}}
While flow-based models offer efficient and deterministic generative modeling, they are fundamentally challenged by \emph{velocity ambiguity} in multi-modal settings. In many real-world robotic tasks, it is common for multiple distinct action trajectories to pass through the same interpolated point $(a_t, t, s)$, typically when $t=0$, each associated with a different underlying velocity $v^* = a_1 - a_0$. In these cases, the flow matching model must assign a unique velocity vector to every $(a_t, t, s)$, even though several valid options exist in the data.
The standard flow matching objective, $\mathcal{L}_\mathrm{FM} = \sum_{i=1}^n \left\| v_\theta(a_t, t, s) - v(a_0^i, a_1^i, t, s) \right\|_2^2,$
requires the model to regress to a single ``best'' velocity at each point. Consequently, the optimal solution is for the model to predict the average of all feasible velocities that pass through the point:
\begin{equation}
    v^*_\theta(a_t, t, s) = \mathbb{E}_{i}\left[ v(a_0^i, a_1^i, t, s) \mid a_t, t, s \right].
\end{equation}
This leads to \emph{mode mixing (averaging effect)}: the predicted velocity may not correspond to any actual expert trajectory, particularly when the data is highly multi-modal. As a result, standard flow matching tends to blur or average over distinct modes, and cannot faithfully represent the true diversity of expert behaviors—highlighting the need for explicit mechanisms to handle multi-modality.

%% file: Sections/4_Method.tex
\begin{figure*}[t]
    \centering
    \includegraphics[width=0.97\textwidth]{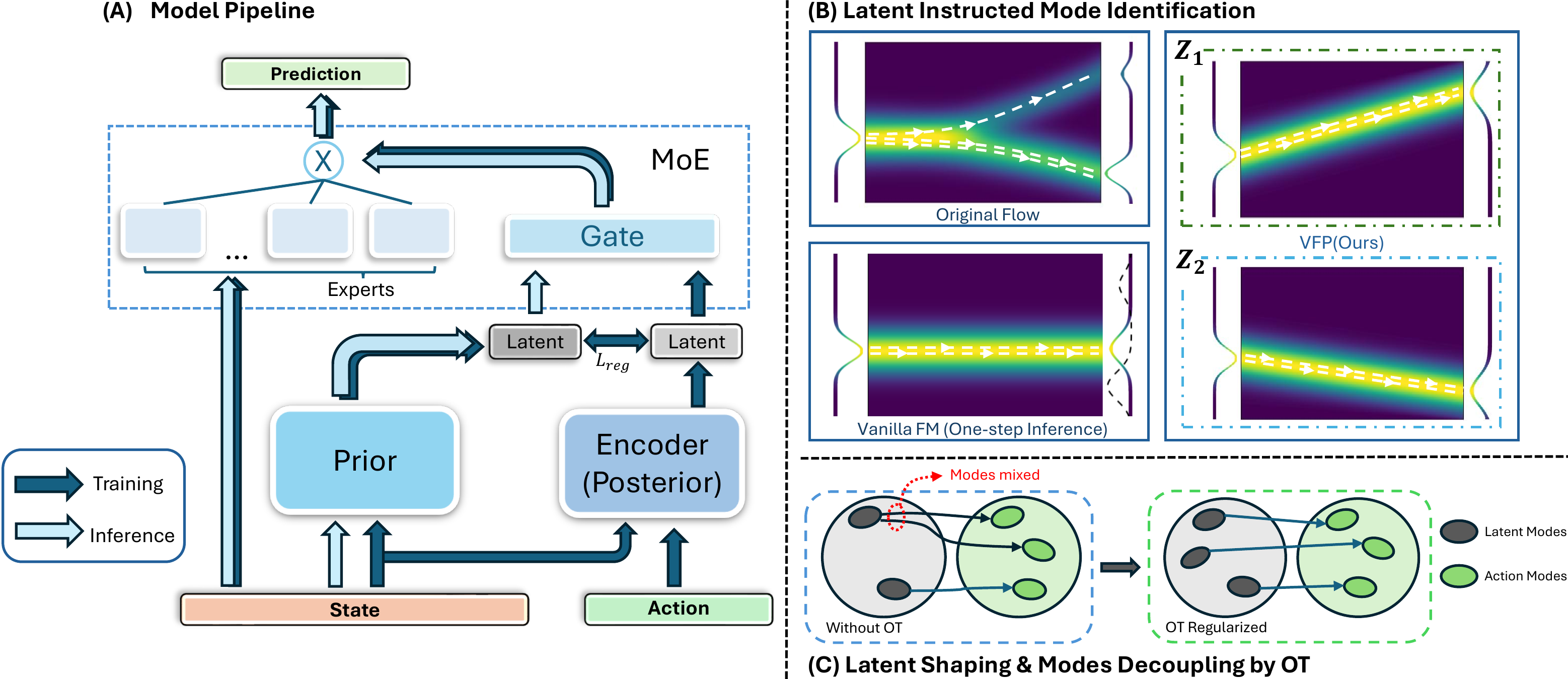}
        \caption{\textbf{Overview of VFP}.
\textbf{(A) Model Pipeline}: The model consists of a latent-conditioned MoE flow matching network. A prior network generates latent variables from input states, which guide the MoE decoder to predict actions efficiently. During training, a posterior encoder is used for variational learning.
\textbf{(B) Latent-Instructed Mode Identification}: Visualization of how VFP enables mode-specific behavior. Compared to the collapsed predictions of vanilla flow matching, our model captures distinct modes conditioned on latent variables.
\textbf{(C) Latent Shaping and Mode Decoupling via OT}: Optimal Transport regularization improves mode separation in the latent space, aligning latent modes with distinct action modes.}
\label{fig:pipeline}
\vspace{-6mm}
\end{figure*}

\section{Method} \label{Sec: Method}

In this section, we present the \emph{Variational Flow Matching Policy} (VFP) framework. VFP employs a variational approach to infer a latent code $z$, which determines the mode selection within the downstream flow matching model (Sec.~\ref{Subsec: VFP structure}). To prevent averaging effect, VFP introduces the static Kantorovich-Optimal Transport (K-OT) regularization, which promotes multi-modal distribution matching in both latent and action distributions throughout the flow matching process (Sec.~\ref{Subsection: KOT}). Furthermore, VFP enhances multi-modal expressiveness by incorporating a mixture-of-experts (MoE) structured flow matching decoder, demonstrating a principled integration of variational inference and MoE for robust policy learning (Sec.~\ref{Subsec: implementation}). An overview of our method is in Fig.~\ref{fig:pipeline}.


\subsection{Variational Flow Matching Structure} \label{Subsec: VFP structure}

In many real-world robot manipulation tasks, the optimal policy $p(a|s)$ is inherently \emph{multi-modal} (MM): given a state $s$, there may exist several distinct, equally valid action choices $a$. Standard learning objectives, such as maximum likelihood or mean-squared error, are insufficient for such settings—they encourage the model to average across modes, producing unrealistic or ambiguous predictions.

\paragraph{\textbf{Multi-Modality Ambiguity Estimation}}
To rigorously measure the magnitude of multi-modality (or ambiguity), we adopt an \emph{ambiguity metric} based on the conditional variance of ground-truth velocities in flow matching. For each state $s$ and interpolated time $t$, consider trajectories that pass through a common intermediate action $a_t = (1-t)a_0 + t a_1$, with $a_0$ and $a_1$ sampled from the data. The oracle velocity for each pair is defined as $v^* = a_1 - a_0$.
The per-point ambiguity at $(a_t, t, s)$ is then quantified by the conditional variance:
\begin{equation}
\begin{aligned}
    \mathrm{Var}[v^* \mid a_t, t, s] & = \mathbb{E}_{(a_0, a_1) | a_t, t, s} \left[ \| v^* - \mu_v \|_2^2 \right], \\
    \text{where} \quad \mu_v & = \mathbb{E}_{(a_0, a_1) | a_t, t, s}[v^*].
\end{aligned}
\end{equation}
This measures the spread of all possible velocities for trajectories passing through the same $(a_t, t, s)$. 
To summarize the overall ambiguity in the data, we define the global ambiguity score by averaging over the data distribution:
\begin{equation}
    \mathcal{A}_\mathrm{FM} = \mathbb{E}_{(a_0, a_1, t, s)} \left[ \mathrm{Var}[v^* \mid a_t, t, s] \right].
\end{equation}
A high value of $\mathcal{A}_\mathrm{FM}$ indicates substantial multi-modality in the expert demonstrations.
However, standard flow matching models are fundamentally limited in handling this ambiguity. When multiple distinct trajectories intersect at the same $(a_t, t, s)$ with different velocities, the model must choose a single velocity for each point. Even with infinite data and perfect optimization, the best the model can do is predict the mean velocity $\mu_v$. Thus, the irreducible loss—i.e., the minimum achievable FM loss—is exactly the ambiguity score $\mathcal{A}_\mathrm{FM}$. This ``crossing term'' is intrinsic: \emph{no deterministic flow matching model can reduce the ambiguity below this floor without an additional mechanism for mode separation}.


\paragraph{\textbf{Variational Flow Matching Policy (VFP) Formulation}}
To address the irreducible ambiguity in standard flow matching, we propose a \emph{Variational Flow Matching Policy} (VFP), which leverages variational inference to capture and disentangle multi-modal structure in the data. As illustrated in Fig.~\ref{fig:pipeline}~(B), the key idea is to introduce a latent variable $z$ that indexes different modes of the action distribution for each state $s$. The overall policy becomes:
\begin{equation}
    p_{\theta, \psi}(a \mid s) = \int p_\theta(a \mid z, s)\, p_\psi(z \mid s)\, dz,
\end{equation}
where $p_\psi(z | s)$ is a learned variational encoder (the “prior”), and $p_\theta(a | z, s)$ is a flow-based decoder.

The flow matching decoder in VFP is constructed similarly to standard FM, but is now conditioned on the latent $z$. For each training example, given $(s, a_0, a_1, t)$ and sampled $z$, we define the interpolated action $a_t = (1-t)a_0 + t a_1$, and train a velocity field $v_\theta(a_t, t, s, z)$ to match the ground-truth displacement $v^* = a_1 - a_0$:
\begin{equation}
    \mathcal{L}_\mathrm{FM} = \mathbb{E}_{s, a_0, a_1, t, z} \left[ \| v_\theta(a_t, t, s, z) - (a_1 - a_0) \|^2 \right].
\end{equation}

For training, we adopt a variational lower bound (ELBO) objective, introducing a recognition network $q_\phi(z \mid a, s)$ to approximate the intractable posterior over $z$. The overall VFP objective is:
\begin{equation}
\begin{aligned}
    \mathcal{L}_\mathrm{VFP} = \mathbb{E}_{(s,a) \sim \mathcal{D},\, z \sim q_\phi(z \mid a, s)}
    \left[
        -\mathcal{L}_\mathrm{FM}(a, s, z) \right. \\
       \left. - D_\mathrm{KL}(q_\phi(z \mid a, s) \| p_\psi(z \mid s))
    \right].
\end{aligned}
\end{equation}
This objective encourages the encoder $p_\psi(z \mid s)$ to capture the multi-modal structure of $p(a|s)$, while the decoder $p_\theta(a \mid z, s)$ learns to model mode-specific, nearly deterministic action flows. By maximizing the ELBO in the context of flow matching, VFP both explains the diversity in the data and regularizes the latent space for generalization.

\paragraph{\textbf{Multi-Modality Ambiguity Decomposition}}
The key theoretical advantage of VFP is that it enables the model to attribute the multi-modality (ambiguity) in the data to the latent variable $z$, thereby simplifying the flow decoder. This can be formalized by decomposing the total ambiguity using the law of total variance. For a fixed $(a_t, t, s)$, the conditional variance of oracle velocities can be written as:
\begin{equation}
\begin{aligned}
\mathrm{Var}[v^* \mid a_t, t, s] = \mathbb{E}_{z \mid a_t, t, s}  \left[ \mathrm{Var}[v^* \mid a_t, t, s, z] \right] &
  \\ 
    + \mathrm{Var}_{z \mid a_t, t, s} \left( \mathbb{E}[v^* \mid a_t, t, s, z] \right) &. 
\end{aligned}
\end{equation}
Here, the first term is the \emph{residual ambiguity} remaining in the decoder after conditioning on $z$, and the second term quantifies the ambiguity explained by the latent variable.
Averaged over the dataset, the global ambiguity score decomposes as:
\begin{equation}
\begin{aligned}
    \mathcal{A}_\mathrm{FM} = \mathcal{A}_\mathrm{VFP} + \mathbb{E}_{(a_t, t, s)} \left[ \mathrm{Var}_{z \mid a_t, t, s}(\mu_v \mid z) \right], \\
    \text{where} \quad \mathcal{A}_\mathrm{VFP} = \mathbb{E}_{(a_0, a_1, t, s)}\left[\mathrm{Var}[v^* \mid a_t, t, s, z]\right].
\end{aligned}
\end{equation}

The second term is always non-negative, yielding the following inequality $ \mathcal{A}_\mathrm{VFP} \leq \mathcal{A}_\mathrm{FM}$.
This proves that variational conditioning strictly reduces or matches the ambiguity present in standard flow matching.

For VFP to be maximally effective, the encoder $p(z \mid s)$ should capture as much of the data's multi-modality as possible—i.e., assign distinct $z$ values to different modes. In the ideal case, where each $z$ uniquely identifies a mode, the residual decoder ambiguity $\mathcal{A}_\mathrm{VFP}$ is minimized, and $p_\theta(a \mid z, s)$ becomes nearly unimodal. Thus, VFP enables the policy to represent highly multi-modal action distributions without forcing the decoder to average over modes, fundamentally overcoming the limitations of standard flow matching in complex, ambiguous settings.


\subsection{Kantorovich-OT for Distribution Matching}
\label{Subsection: KOT}

In multi-modal imitation learning, the central challenge is to match the \emph{entire} conditional action distribution $p(a \mid s)$ induced by the expert, not just individual demonstration pairs. For any given state $s$, the expert policy $p_{\mathrm{expert}}(a \mid s)$ may assign probability mass to multiple diverse actions $a$, reflecting the inherent ambiguity and richness of real-world decision-making. Conventional flow matching and sample-based imitation approaches focus on aligning isolated samples or trajectories, i.e., matching $(s, a)$ pairs. However, such pointwise objectives are fundamentally limited: they often induce mode averaging, resulting in learned policies $p_\theta(a \mid s)$ that underrepresent the true diversity of expert behaviors. To faithfully imitate expert demonstrations in multi-modal settings, it is essential to align distributions at the population level—ensuring that $p_\theta(a \mid s)$ covers all modes present in $p_{\mathrm{expert}}(a \mid s)$ for each state $s$.

\paragraph{\textbf{Kantorovich-Optimal Transport (K-OT) Formulation}}
The K-OT framework provides a principled approach to aligning probability distributions by minimizing the cost of transporting probability mass from one distribution to another. In the setting of multi-modal imitation learning, for each state $s$, we consider two discrete distributions: the predicted action distribution $p_\theta(a \mid s) = \sum_{i} u_i\,\delta_{a_i}$ (where $a_i$ are samples from the policy) and the empirical expert action distribution $p_{\mathrm{expert}}(a \mid s) = \sum_{j} v_j\,\delta_{a'_j}$ (where $a'_j$ are expert demonstrations). Here, $u_i$ and $v_j$ are normalized weights summing to one, and $\delta_{a}$ denotes a Dirac delta at $a$.

The K-OT distance between $p_\theta(a \mid s)$ and $p_{\mathrm{expert}}(a \mid s)$ is defined as the minimum total transport cost under a ground cost $\mathcal{C}(a, a')$:
\begin{equation} \label{Eqn: OT}
    \mathrm{OT}(p_\theta, p_{\mathrm{expert}}) = \min_{\gamma \in \Pi(u, v)} \sum_{i, j} \gamma_{i, j} \, \mathcal{C}(a_i, a'_j),
\end{equation}
where $\gamma \in \Pi(u, v)$ denotes the set of all valid couplings (transport plans) between the two distributions with marginals $u$ and $v$.
Typically, the ground cost is chosen as the squared Euclidean distance $\mathcal{C}(a, a') = \|a - a'\|^2$.
This formulation encourages each predicted action to be matched with a similar expert action, while the optimal coupling $\gamma$ ensures the entire distribution is faithfully aligned. In practice, since states are continuous, we define the OT cost as $\|a - a'\|^2 + \lambda \|s - s'\|^2$ to enable soft cross-condition matching without exact state alignment. We adopt Sinkhorn optimal transport to compute a differentiable form of this regularization.

\paragraph{\textbf{K-OT Regularization}}
To promote faithful multi-modal imitation, we introduce Kantorovich-Optimal Transport (K-OT) as a regularization term in the training objective. For each mini-batch, we sample sets of actions from the policy $p_\theta(a \mid s)$ and the expert demonstrations $p_{\mathrm{expert}}(a \mid s)$ for each state $s$. The K-OT distance between these sets is then computed and used to regularize the learning process:
\begin{equation} \label{Eqn: OT loss}
    \mathcal{L}_\text{total} = \mathcal{L}_\mathrm{VFP} + \alpha\,\mathbb{E}_s\left[\mathrm{OT}\left( p_\theta(a \mid s),\, p_{\mathrm{expert}}(a \mid s) \right)\right],
\end{equation}
where $\alpha$ is a weighting coefficient controlling the influence of K-OT regularization.
This approach ensures that, beyond trajectory-level imitation via variational flow matching, the model is explicitly encouraged to align the entire predicted action distribution with the expert's for each state. Therefore, as depicted in Fig.~\ref{fig:pipeline}~(C), K-OT regularization helps the policy cover all demonstrated modes, mitigates averaging effect, and produces more diverse and expressive behaviors that better reflect the richness of expert demonstrations.




\subsection{VFP Implementation}
\label{Subsec: implementation}
To further enhance the model's capacity for multi-modality and accelerate both training and inference, as shown in Fig.~\ref{fig:pipeline}~(A), we implement the policy decoder as a \textbf{Mixture-of-Experts} (MoE). The MoE structure enables each expert to specialize in a particular behavioral mode, while the gating network efficiently selects and combines expert predictions. This architecture not only improves expressiveness for complex, multi-modal action distributions, but also allows for parallelized computation and scalable learning.


In our VFP framework, multi-modality is captured in the latent variable $z$, which is sampled from a diffusion-based prior $p_\psi(z \mid s)$. The decoder is a Mixture-of-Experts (MoE) flow model, where the velocity field consists of $K$ flow matching experts. Each expert $v_{\theta,i}(a_t, t, s)$ models a distinct behavioral mode, and a gating network maps $z$ to a set of coefficients $\{g_i(z)\}_{i=1}^K$ satisfying $\sum_{i=1}^K g_i(z) = 1$.
Therefore, during training, we use a decoupled loss formulation where each expert is trained independently, and the total loss is computed as a gating-weighted sum:
\begin{equation}
    \mathcal{L}_{\mathrm{VFP}} = \mathcal{L}_{\mathrm{MoE}} = \sum_{i=1}^K g_i(z) \cdot \left\| v_{\theta,i}(a_t, t, s) -v \right\|^2.
\end{equation}
This encourages expert specialization, where each expert independently fit a mode. Accordingly, during inference, we select the expert with the highest gating weight, i.e., $i^* = \arg\max_i g_i(z)$, and generate actions using only $v_{\theta,i^*}$.

To summarize, the overall VFP framework leverages a diffusion-based latent prior and a Mixture-of-Experts flow decoder to enable robust and expressive multi-modal policy learning. By integrating variational inference, K-OT regularization, and expert specialization, VFP achieves efficient distribution-level alignment and adaptive, diverse robot behavior in complex manipulation tasks.

%% file: Sections/5_Experiment.tex
\section{Experiments} \label{Sec: experiments}

We evaluate the VFP framework on a suite of challenging, multi-modal robot manipulation tasks. Our experiments are designed to answer the following core questions:
\begin{itemize}
    \item \textbf{Effectiveness:} Does VFP improve manipulation performance in multi-modal settings compared to baseline methods, and does this advantage transfer to real-robot settings?
    \item \textbf{Efficiency:}  Can VFP maintain the fast inference advantage of traditional flow-matching models? Does it require more computational resources?
    \item \textbf{Ablation:} How important are K-OT regularization and the Mixture-of-Experts (MoE) flow decoder for achieving state-of-the-art results?
\end{itemize}
The remainder of this section describes our experimental setup, baselines, evaluation protocols, and results.






\subsection{Experiment Setting}

\begin{figure}[t]
    \centering
    \includegraphics[width=0.97\columnwidth]{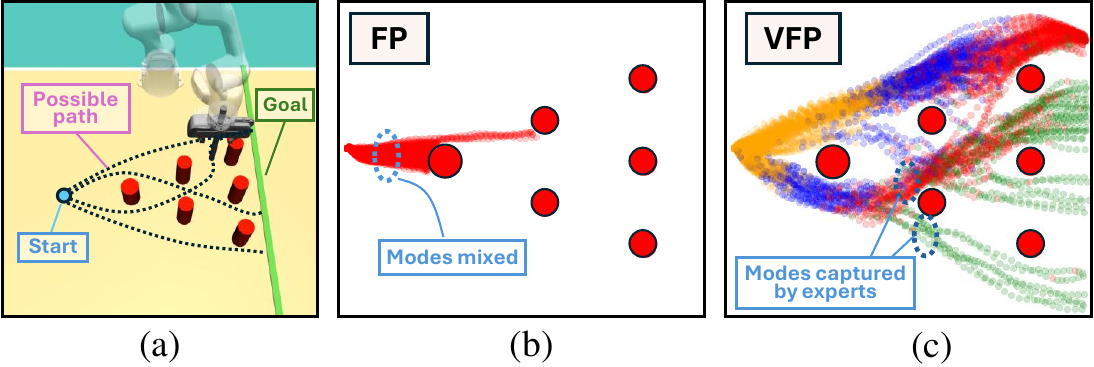}
    \caption{The Avoiding task and behaviors of policies. \textbf{(a)}: The environment of avoiding. \textbf{(b)}: Behavior of FlowPolicy (red trajectories). \textbf{(c)}: Behavior of VFP (ours). Movements made by different experts are in different colors.}
    \label{Fig: avoidingTrajs}
\end{figure}

\begin{figure}[t]
  \centering
  \begin{subfigure}[t]{0.49\linewidth}
    \centering
    \includegraphics[width=\linewidth]{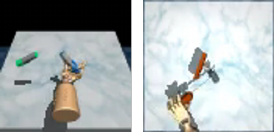}
    \caption{Adroit: Pen and Hammer}
    \label{fig:teasersub1}
  \end{subfigure}%
  \hfill
  \begin{subfigure}[t]{0.49\linewidth}
    \centering
    \includegraphics[width=\linewidth]{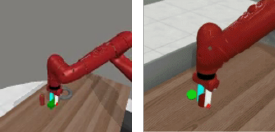}
    \caption{MW: Assembly and Push}
    \label{fig:teaserSub2}
  \end{subfigure}
  \vspace{-1mm}
  \caption{Tasks in (a) Adroit and (b) Meta-World.}
  \label{fig:adroitMeta}
\vspace{-4mm}
\end{figure}

\paragraph{\textbf{Environments}}
In simulated experiments, we evaluate VFP in two categories of multi-modal robot manipulation environments: task and path multi-modality. For task multi-modality, where policies should commit to a specific objective instead of mixing multiple tasks, we use the Franka Kitchen environment~\cite{frankaKitchen} (shown in Fig.~\ref{Fig: teaser} (a)), which requires the robot to complete any subset of seven distinct kitchen tasks in arbitrary order. For path multi-modality, where the policy must distinguish between different valid trajectories rather than averaging them, we consider the D3IL obstacle avoidance task (shown in Fig.~\ref{Fig: avoidingTrajs}) as a clear example, since the robot can reach its goal by avoiding obstacles via multiple paths. We also evaluate on two large-scale environments, Adroit~\cite{adroit} and Meta-World~\cite{metaworld} (shown in Fig.~\ref{fig:adroitMeta}), for broader evaluations.

For real-world settings, we consider a real-world variant of \emph{Avoiding} task and two extra bimanual tasks of rich multi-modality: \emph{Tubes Placement} and \emph{Cups Nesting}. For each task, we collect 10 trajectories via teleoperation.


\paragraph{\textbf{Baselines and Evaluation Metrics}}
We compare VFP to state-of-the-art diffusion- and flow-based imitation learning methods, including Diffusion Policy (DP)~\cite{diffplcy}, 3D Diffusion Policy (DP3)~\cite{3ddiff}, and FlowPolicy (FP)~\cite{flowpolicy}. These baselines are chosen for their strong performance on multi-modal robot manipulation benchmarks. To evaluate effectiveness, we report in simulation the success rate averaged over multiple seeds and episodes (mean ± std), and on real robots the success counts out of 10 trials. For efficiency, we measure inference time, the number of function evaluations (NFE), and computational resource usage, quantified by the number of active NN parameters used during policy inference.

\begin{table*}[t]
    \centering
    \caption{Comparison of success rates (\%) $\uparrow$ on Adroit and Meta-World}
    
    \resizebox{0.98\textwidth}{!}{
    \begin{threeparttable}
    \begin{tabular}{l|c|ccc|cccc|c}
    \toprule
    \multirow{2}{*}{Methods} & \multirow{2}{*}{NFE} & \multicolumn{3}{c|}{Adroit (\%)}  & \multicolumn{4}{c|}{Meta-World (\%)} & \multirow{2}{*}{\textbf{Average (\%)}} \\
     &  & Hammer & Door & Pen & Easy(21) & Medium(4) & Hard(4) & Very Hard(5) & \\  \midrule
    DP~\cite{diffplcy} & 10 &16$\pm${10}&34$\pm${11}&13$\pm${2}& 50.7$\pm${6.1} & 11.0$\pm${2.5}& 5.25$\pm${2.5}& 22.0$\pm${5.0}& 35.2$\pm${5.3} \\
    DP3~\cite{3ddiff} & 10 &100$\pm${0}&56$\pm${5}&46$\pm${10}& 87.3$\pm${2.2} & 44.5$\pm${8.7}& 32.7$\pm${7.7}& 39.4$\pm${9.0}& 68.7$\pm${4.7} \\
    FlowPolicy~\cite{flowpolicy}  & 1  &\textbf{100$\pm${0}} &58$\pm${5}& 53$\pm${12}& 90.2$\pm${2.8}& 47.5$\pm${7.7}&37.2$\pm${7.2}& 36.6$\pm${7.6}& 70.0$\pm${4.7} \\ 
    \midrule
    \textbf{VFP (Ours)}             & 1  &99$\pm${1} &\textbf{64$\pm${3}}& \textbf{61$\pm${8}}& \textbf{91.5$\pm${1.2}}& \textbf{48.4$\pm${8.1}}&\textbf{37.8$\pm${8.2}}& \textbf{40.5$\pm${6.7}}& \textbf{72.8}$\pm${\textbf{3.7}} \\
    VFP w/o K-OT & 1  &96$\pm${3} &61$\pm${6}& 54$\pm$13
    & 90.8$\pm${1.7}& 48.2$\pm${8.8}&36.5$\pm${9.1}& 36.2$\pm${7.2}& 71.3$\pm$4.4
    \\
    VFP w/o MoE  & 1  &94$\pm${2} &57$\pm${6}& 50$\pm${11}
    & 88.2$\pm${4.8}& 46.6$\pm${9.2}&35.5$\pm${5.2}& 35.2$\pm${4.6}& 69.1$\pm${5.4}
    \\
    \bottomrule
    \end{tabular}

\end{threeparttable}
}

\label{tab:adroit_metaworld_results}
\vspace{-2mm}
\end{table*}

\begin{table}[t]
\centering
\setlength{\tabcolsep}{1.2mm}

\caption{Success Rate/Reward on D3IL and Franka Kitchen}
\begin{tabular}{lccc|c}
\toprule
\multirow{2}{*}{Methods} & \multicolumn{3}{c|}{D3IL (\%)} & \multirow{2}{*}{Kitchen} \\
& Avoiding & Sorting-4 & Pushing & \\
\midrule
DP & 0.89$\pm$0.01 & 0.30$\pm$0.06 & 0.92$\pm$0.03 & 3.99 \\
FlowPolicy & 0.02$\pm$0.01 & 0.14$\pm$0.01 & 0.67$\pm$0.02 & 3.64 \\
VFP (Ours)  & \textbf{0.91$\pm$0.01} & \textbf{0.45$\pm$0.01} & \textbf{0.94$\pm$0.02} & \textbf{4.06} \\
\bottomrule
\end{tabular}
\label{tab:d3ilFranka}
\vspace{-2mm}
\end{table}

\subsection{Simulated Experiments}


\paragraph{\textbf{VFP achieves task multi-modality by latent mode selection}}
We evaluate our method on the Franka Kitchen benchmark, which requires the agent to select and complete multiple tasks within a single environment. In the experiments, we observe that FlowPolicy often oscillates between tasks—such as switching repeatedly among the three light switches in the turn-on-light task—and this indecision also leads to frequent failures in handle grasping tasks. In contrast, our VFP leverages the latent prior to effectively distinguish between task modes, enabling the flow-matching decoder to generate consistent and smooth trajectories for each selected task. As a result, VFP increases the average rewards in Franka Kitchen by $11.5\%$ (Tab.~\ref{tab:d3ilFranka}), compared to FlowPolicy. Also, in the Pushing task from D3IL, our success increases by $40.2\%$.

\paragraph{\textbf{VFP captures path multi-modality by disentangling trajectory modes}}
We evaluate path-level multi-modality on the D3IL benchmark, where each task permits multiple valid trajectories to reach the goal. As shown in Tab.~\ref{tab:d3ilFranka}, VFP significantly outperforms baseline methods across all three tasks. As shown in Fig.~\ref{Fig: avoidingTrajs}, FlowPolicy frequently exhibits averaging behavior, causing the agent to move directly into obstacles instead of selecting a valid path around them. In contrast, VFP effectively captures the multi-modal structure of the demonstrations, enabling the agent to reliably choose distinct trajectories. It's also shown in Fig.~\ref{Fig: avoidingTrajs} that our model produces diverse behaviors in rollouts, with different experts specializing in different trajectory modes.






We also evaluate VFP on the large-scale Adroit and Meta-World manipulation environments. While multi-modality in these benchmarks is generally less pronounced than in D3IL, some complex tasks—such as swinging a hammer or rotating a pen—feature various possible trajectories and thus introduce path-level multi-modality. Overall, VFP improves average performance across all tasks by about $4\%$ compared to FlowPolicy, yielding comparable results. However, for more challenging tasks that require greater precision and control, VFP provides more substantial gains. For example, in the Adroit \emph{Pen} task, where the agent must use dexterous hand movements to rotate a pen, FlowPolicy often falls back on crude wrist oscillations instead of the fine finger manipulations needed for successful rotation. This is likely due to mode mixing, where distinct rotation strategies become entangled in the model’s output distribution, preventing the policy from committing to an effective trajectory. In contrast, VFP avoids this issue and increases the average success rate by $15\%$ on the \emph{Pen} task. Similarly, for ``Very Hard'' tasks in Meta-World, VFP achieves a $10\%$ improvement over FlowPolicy.

\begin{table}[t]
\centering
\setlength{\tabcolsep}{1mm}
\caption{Comparison on inference time and model size}
\label{tab:efficiency}

\resizebox{0.98\columnwidth}{!}{
\begin{threeparttable}
\begin{tabular}{l|cccc|cc}
\toprule
& \multicolumn{4}{c|}{Inference time (ms) $\downarrow$}  & \multicolumn{2}{c}{Active params (M) $\downarrow$} \\
Methods & D3IL & Adroit & Metaworld & \textbf{Avg.} & D3IL & Adroit/MW \\
\midrule
DP   & 56.2 & 100.3 & 106.5 & 87.7 & \textbf{0.59} & 255.1  \\
DP3  & -- & 145.9 & 145.6 & -- & -- & 255.2  \\
FP   & \textbf{13.1} & \textbf{20.1} & \textbf{19.9} & \textbf{17.7} & \textbf{0.59} & 255.7  \\
VFP& 14.3& 21.5& 20.3& 18.7 & 0.60 & \textbf{243.3}  \\
\bottomrule
\end{tabular}
\begin{tablenotes}
\item  “--” indicates that DP3 is not applicable to D3IL.
\end{tablenotes}
\end{threeparttable}
}
\vspace{-3mm}
\end{table}




\paragraph{\textbf{Model Efficiency}}
We evaluate the average inference time and model size on D3IL, Adroit, and Meta-World. As shown in Tab.~\ref{tab:efficiency}, our method incurs only a $5.6\%$ increase in inference time compared to FlowPolicy, while remaining $4.6\times$ faster than diffusion-based models. The active model size remains comparable overall, and is even $4.9\%$ smaller on Adroit and Meta-World.
Although VFP incorporates additional components and achieves substantially improved performance, it still maintains high computational efficiency. This is primarily due to the Mixture-of-Experts (MoE) decoder, which decomposes the overall task into smaller subproblems handled by individual experts. Each expert operates with a shallower network, thereby reducing the per-expert computational cost and keeping the overall inference overhead low.

\begin{figure*}[t]
    \centering
    \includegraphics[width=0.97\textwidth]{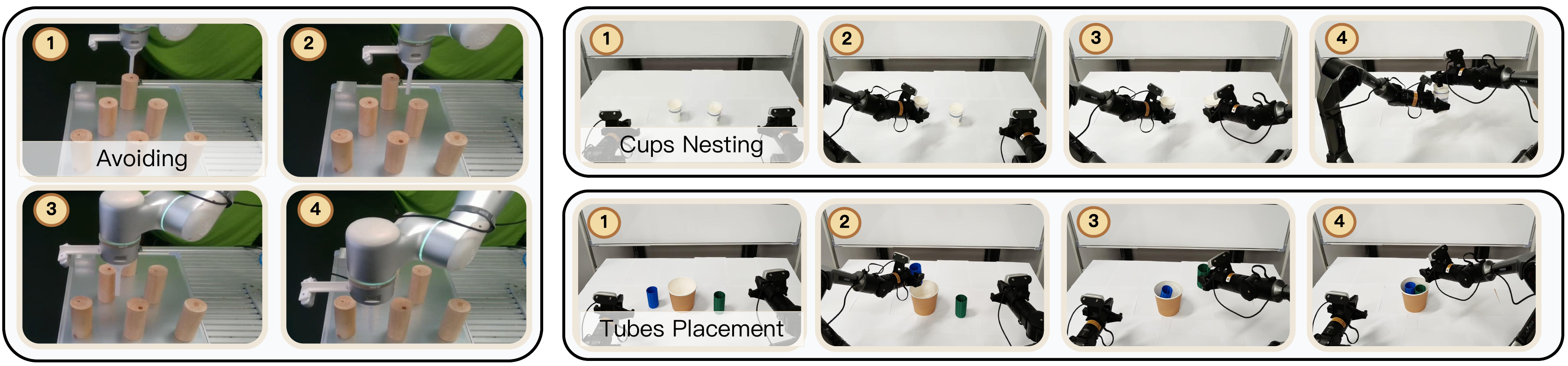}
        \caption{Experiments on Real-World \emph{Avoiding}, \emph{Cups Nesting} and \emph{Tubes Placement}.}
\label{fig:real_bimanual_results}
\vspace{-3mm}
\end{figure*}

\subsection{Ablation Studies}

In our ablation studies, we aim to demonstrate how VFP benefits from the OT regularization and the MoE-structured decoder in enhancing its ability to model multi-modal distributions. To this end, we primarily conduct ablation studies on D3IL and Franka Kitchen, two benchmarks that exhibit rich multi-modality. Additionally, the ablated variants, VFP w/o K-OT and VFP w/o MoE, are also evaluated on Adroit and Meta-World to assess their generalization in broader settings.  

\begin{figure}[t]
  \centering
  \includegraphics[width=0.98\columnwidth]{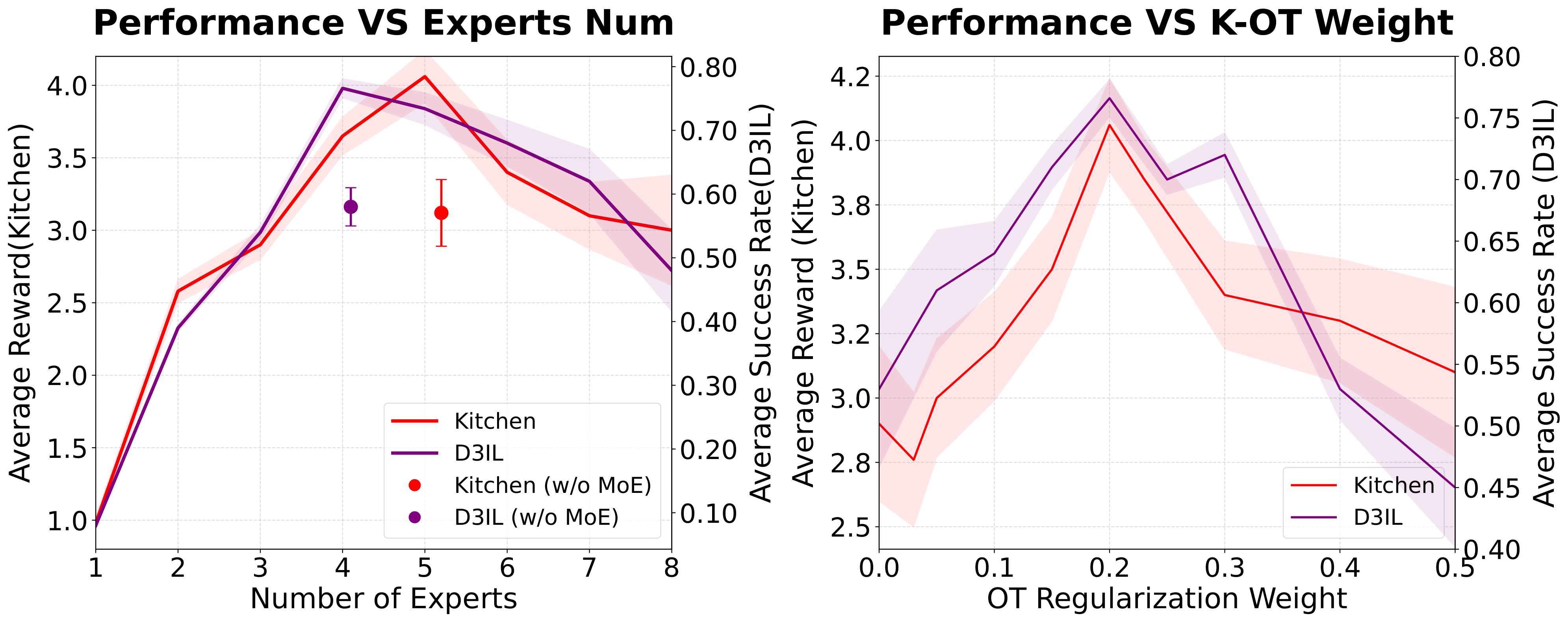}
  \caption{Ablation studies. 
\textbf{Left:} Performance under varying numbers of experts in the MoE decoder. Circle markers denote single-model variants without MoE, placed at x-axis positions indicating the number of experts with roughly equivalent total parameter count.
\textbf{Right:} Performance under different OT regularization weights.  }
  \label{fig:abaltionPlots}
\vspace{-3mm}
\end{figure}

\paragraph{\textbf{K-OT Regularization}}
To investigate the effect of OT regularization, we conduct ablation studies by varying the weight of the OT loss. As reported in Tab.~\ref{tab:adroit_metaworld_results}, in environments with less pronounced multi-modality (Adroit and Meta-World), OT regularization yields a modest improvement of around $2\%$. In contrast, as shown in Fig.~\ref{fig:abaltionPlots}, in highly multi-modal benchmarks such as D3IL and Franka Kitchen, properly tuning the OT weight leads to performance gains of $44\%$ and $36\%$, respectively, compared to models trained without OT.
Moreover, the standard deviation of success rate (or reward) across different seeds decreases by $27\%$ on D3IL and $18\%$ on Meta-World, indicating improved training stability and more consistent multi-modal behavior. These results highlight that OT regularization is particularly beneficial in multi-modal environments, enhancing both overall performance and the robustness of the learning process.

\paragraph{\textbf{MoE Decoder}}
To evaluate the importance of the MoE structure, we conduct experiments using a variant where the entire MoE decoder is replaced by a single monolithic neural network, directly conditioned on the latent variable and scaled to have a similar total number of parameters as the original MoE-based model. As shown in Tab.~\ref{tab:adroit_metaworld_results} and Fig.~\ref{fig:abaltionPlots}, while the MoE decoder yields a moderate average improvement of approximately $5.3\%$ on Adroit and Meta-World, it leads to significantly larger gains on D3IL and Franka Kitchen—around $26\%$ and $32\%$, respectively. These results demonstrate that, under similar model capacity, the MoE structure substantially enhances the ability to model multi-modal distributions by dividing the problem space of complex, multi-modal distributions.
Additionally, to provide practical guidance on selecting the number of experts, we conduct further ablation studies. As shown in Fig.~\ref{fig:abaltionPlots}, increasing the number of experts initially improves performance by expanding model capacity; however, an excessively large number leads to degraded performance and training instability, likely due to the limited capacity of the gating network to effectively assign expert responsibilities.
\begin{table}[t]
\centering
\caption{Success counts in real-world experiments}
\resizebox{0.7\columnwidth}{!}{%
\begin{tabular}{lccc}
\toprule
Methods & Avoiding & Cups & Tubes \\
\midrule
DP          & 6/10 & 2/10 & 2/10 \\
FlowPolicy  & 0/10 & 0/10 & 1/10 \\
VFP (Ours)  & \textbf{8/10} & \textbf{3/10} & \textbf{4/10} \\
\bottomrule
\end{tabular}
}
\label{tab:real-world}
\vspace{-5mm}
\end{table}


\subsection{Real-Robot Experiments}
To examine our conclusions in real-world settings, we evaluate three hardware tasks: a real-world variant of \emph{Avoiding} and two bimanual tasks—\emph{Cups Nesting} and \emph{Tubes Placement} (Fig.~\ref{fig:real_bimanual_results}). 
\emph{Cups Nesting} requires picking two cups and nesting one into the other; either cup may go on top. 
\emph{Tubes Placement} requires grasping two tubes and placing them into a bucket; the placement order may vary.

Across all three tasks, behaviors mirror our simulation findings. FlowPolicy fails to resolve multi-modality: when multiple valid choices exist, it averages across modes and produces indecisive, symmetric motions (e.g., both hands rising to similar heights and moving toward the midline in \emph{Cups Nesting}), leading to cups collisions or drops. In contrast, VFP separates modes and commits early to a single hypothesis, yielding coherent, role-assigned trajectories. Quantitatively, as reported in Tab.~\ref{tab:real-world}, on \emph{Avoiding}, while FlowPolicy still shows constant failures, VFP succeeds in 8/10 trials (also exceeding DP by 33\%). Also, VFP attains 3/10 and 4/10 successes on \emph{Cups Nesting} and \emph{Tubes Placement}, respectively, compared with FlowPolicy’s 0/10 and 1/10 (with DP at 2/10 for both). The smooth motion enabled by fast inference further contributes to VFP’s advantage over DP on hardware.

We observe that the performance gap between VFP and FlowPolicy is even larger on real robots than in simulation. We attribute this to the heightened demand for genuine multi-modal competence on hardware, where averaged, indecisive actions are less tolerable and lead to contact errors. This underscores \textsc{VFP}’s ability to capture multi-modality on hardware and indicates that, in real-world settings, \textsc{VFP} is not merely a superior alternative to standard flow-matching policies, but a necessary choice for practical deployment.

%% file: Sections/6_Conclusion.tex
\section{Conclusion} \label{Sec: conclusion}

We propose Variational Flow-Matching Policy (VFP), a method designed to effectively address multi-modality in flow-based imitation learning for robot manipulation. By leveraging a variational latent prior, Kantorovich Optimal Transport regularization, and a Mixture-of-Experts (MoE) decoder, VFP enhances both multi-modal representation and computational efficiency. Simulated benchmarks confirm that VFP captures multi-modality and achieves significant success-rate improvements without sacrificing inference speed. Real-world results further emphasize the necessity of VFP for practical deployment. Ablation studies further confirm the effectiveness of both K-OT regularization and the MoE decoder as key components of our framework.